\documentclass[sigconf]{acmart}

\setcopyright{none}
\renewcommand\footnotetextcopyrightpermission[1]{}

\AtBeginDocument{%
  }

\begin{document}

\title{CUAAudit: Meta-Evaluation of Vision-Language Models as Auditors of Autonomous Computer-Use Agents}

\author{Marta Sumyk}
\email{sumyk.pn@ucu.edu.ua}
\affiliation{
  \institution{Ukrainian Catholic University}
  \city{Lviv}
  \country{Ukraine}
}

\author{Oleksandr Kosovan}
\email{o.kosovan@ucu.edu.ua}
\affiliation{
  \institution{Ukrainian Catholic University}
  \city{Lviv}
  \country{Ukraine}
}

\begin{abstract}
Computer-Use Agents (CUAs) are emerging as a new paradigm in human-computer interaction, enabling autonomous execution of tasks in desktop environment by perceiving high-level natural-language instructions. As such agents become increasingly capable and are deployed across diverse desktop environments, evaluating their behavior in a scalable and reliable manner becomes a critical challenge. Existing evaluation pipelines rely on static benchmarks, rule-based success checks, or manual inspection, which are brittle, costly, and poorly aligned with real-world usage. In this work, we study Vision-Language Models (VLMs) as autonomous auditors for assessing CUA task completion directly from observable interactions and conduct a large-scale meta-evaluation of five VLMs that judge task success given a natural-language instruction and the final environment state. Our evaluation spans three widely used CUA benchmarks across macOS, Windows, and Linux environments and analyzes auditor behavior along three complementary dimensions: accuracy, calibration of confidence estimates, and inter-model agreement. We find that while state-of-the-art VLMs achieve strong accuracy and calibration, all auditors exhibit notable performance degradation in more complex or heterogeneous environments, and even high-performing models show significant disagreement in their judgments. These results expose fundamental limitations of current model-based auditing approaches and highlight the need to explicitly account for evaluator reliability, uncertainty, and variance when deploying autonomous CUAs in real-world settings.
\end{abstract}





\maketitle

\begingroup
\renewcommand\thefootnote{}% empty number style
\footnotetext{This work has been accepted to appear at the HEAL @ CHI 2026 Worshop on
Human-centered Evaluation and
Auditing of Language Models.}
\addtocounter{footnote}{1} % so the next footnote starts from 1
\endgroup

\section{Introduction}

Recent advances in large language models and multimodal perception have given rise to Computer-Use Agents (CUAs): autonomous systems that can operate Graphical User Interfaces (GUIs) by translating high-level natural-language instructions into sequences of actions such as clicking, typing, scrolling, and dragging \cite{sager2025comprehensivesurveyagentscomputer}.

From a Human–Computer Interaction (HCI) perspective, CUAs extend a long line of work on interface agents and intelligent user interfaces, where users attribute intent, agency, and social meaning to interactive systems rather than viewing them as purely functional tools \cite{10.1145/1314161.1314180}. Recent systems further demonstrate that large vision-language models can act as unified controllers for complex desktop environments, generalizing across applications, tasks, and operating systems without relying on handcrafted rules \cite{wang2025uitars2technicalreportadvancing}. As a result, CUAs offer a service-agnostic alternative to traditional robotic process automation, reducing brittleness and maintenance costs while supporting a broader range of real-world tasks \cite{sager2025comprehensivesurveyagentscomputer}.

Beyond automation, CUAs hold particular promise for accessibility and inclusive interaction. When paired with natural-language or voice interfaces, they enable users with motor, visual, or cognitive impairments to complete multi-step tasks through language alone \cite{vu2024gptvoicetaskeradvancingmultistepmobile, zhang2025largelanguagemodelbrainedgui}. More broadly, CUAs can reduce cognitive and interaction burdens for non-technical users, older adults, and individuals facing language or executive-function challenges \cite{bigham2020assistivetechnology}.

As CUAs are increasingly deployed in real-world settings, rigorous evaluation prior to deployment becomes essential. However, assessing CUA behavior remains a fundamental challenge. Existing evaluation pipelines rely on static benchmarks, rule-based success checks, or manual inspection, all of which are costly to maintain, brittle to interface changes, and poorly aligned with real-world usage \cite{xie2024osworldbenchmarkingmultimodalagents}. Such approaches typically yield coarse success signals and provide limited insight into partial task completion, user-acceptable failures, or performance under realistic UI variation. These limitations are especially concerning given that CUAs act autonomously on users’ behalf, often across multiple applications and involving sensitive data.

In this work, we study Vision-Language Models (VLMs) as autonomous auditors for CUAs. Rather than relying on internal agent states or handcrafted evaluation logic, VLM-based auditors assess task completion directly from observable evidence by judging whether a natural-language instruction has been satisfied in the final GUI state. We conduct a large-scale meta-evaluation of VLM auditors across multiple operating systems and benchmarks, analyzing their accuracy, confidence calibration, and inter-model agreement. By treating evaluation as a first-class problem, our study characterizes the reliability and limitations of model-based auditing and highlights key challenges for the safe and robust deployment of CUAs in real-world settings.

\section{Related Works}

\subsection{Computer-Use Agents and GUI Automation}

Research on CUAs builds on a long history of GUI automation, robotic process automation (RPA), and intelligent user interfaces. Early systems relied on handcrafted rules, application-specific scripts, DOM trees, or accessibility APIs to automate repetitive tasks. While effective in controlled settings, these approaches were brittle to interface changes, required substantial manual maintenance, and failed to generalize across applications or operating systems \cite{bigham2020assistivetechnology}.

Recent work has shifted toward learning-based approaches that operate directly on multimodal observations of the interface, typically combining screenshots with natural-language task instructions. This paradigm enables agents to interact with graphical user interfaces through the same perceptual and control channels available to human users. Systems such as SeeAct \cite{zheng2024gpt4visiongeneralistwebagent}, InfiGUIAgent \cite{liu2025infiguiagentmultimodalgeneralistgui}, SEAGENT \cite{sun2025seagentselfevolvingcomputeruse}, and UI-TARS \cite{wang2025uitars2technicalreportadvancing} demonstrate that large vision-language models can act as general-purpose GUI controllers in a wide range of desktop and mobile environments.

Collectively, these results show that CUAs can achieve substantial cross-application and cross-platform generalization without relying on application-specific APIs or predefined workflows. By treating the GUI as an executable visual environment rather than a structured programmatic interface, CUAs represent a departure from traditional automation pipelines and enable more flexible, service-agnostic interaction with existing software ecosystems.

\subsection{CUA as a New HCI Concept}

CUAs introduce an emerging interaction paradigm in which users delegate high-level goals to autonomous agents that perceive, reason, and act directly within existing GUIs. Unlike traditional interaction models based on direct manipulation \cite{shneiderman1983direct}, CUAs function as intermediaries that execute tasks on the user’s behalf through the same visual and control channels available to humans.

From an HCI perspective, CUAs build upon earlier work on interface agents and intelligent user interfaces, which explored how software agents could assist users through recommendations, reminders, or adaptive behavior \cite{maes1994agents, lieberman1997autonomous}. These systems, however, typically played a supportive or advisory role and relied on structured application access, predefined workflows, or handcrafted rules. In contrast, modern CUAs are designed for end-to-end task execution: given a natural-language instruction, the agent must interpret user intent, observe the current interface state, plan a sequence of actions, and adapt its behavior in dynamic and partially observable environments.

This shift places CUAs within the tradition of mixed-initiative interaction and human--automation collaboration, where control is shared between humans and autonomous systems \cite{LIU2025237, horvitz1999principles, hearst1999mixed}. However, CUAs push this paradigm further by substantially reducing direct user oversight during task execution. The graphical user interface becomes an executable environment rather than a passive display, and interaction is reframed as a sequential decision-making process over perceptual inputs and actions such as clicking, typing, scrolling, or dragging. This framing aligns CUAs with agent-based models of perception--action loops in interactive systems \cite{russell2010artificial}.

At the same time, increased autonomy introduces challenges central to HCI research on trust, safety, and usability. Prior work shows that reduced human control can lead to loss of transparency, over-reliance on automation, and difficulty diagnosing or recovering from failures \cite{lee2004trust, norman1990automation}. Because CUAs act directly on users’ behalf, often across multiple applications and involving sensitive data, misaligned or unsafe behavior may have immediate and costly consequences, amplifying the need for reliable evaluation and auditing mechanisms.

\subsection{Agents Audit}
As autonomous agents are increasingly deployed in real-world settings, systematically auditing their behavior has become a central concern. Agent auditing broadly refers to evaluating correctness, reliability, safety, and alignment with intended objectives, particularly in sequential and interactive environments \cite{amodei2016concrete, doshi2017towards}.

Traditional agent evaluation has focused on structured environments such as simulators or benchmarks with explicit reward functions or success criteria. Related work on verification and testing explores formal methods, constraint checking, and adversarial stress testing, but similarly relies on structured state representations and predefined safety properties \cite{katz2017reluplex}. These assumptions often break down in open-ended, real-world interfaces.

With the rise of large language models and tool-using agents, recent work has explored evaluation under less structured conditions using human judgment, preference learning, or learned reward models \cite{christiano2017deep, ouyang2022training}. While effective in some contexts, these approaches often require human-in-the-loop supervision or access to agent internals, limiting scalability and applicability to complex GUI-based environments.

More recently, a small number of studies have begun to examine autonomous evaluation of CUAs \cite{lin2025cuarewardbenchbenchmarkevaluatingreward, sumyk2025areyetvisionbasedjudge}. These works demonstrate the feasibility of model-based evaluators in realistic desktop settings, but remain limited in scope—typically focusing on a narrow set of tasks, metrics, or operating systems. As a result, key challenges such as cross-platform generalization, evaluator reliability, and robustness under diverse interaction patterns remain underexplored.

Overall, CUAs expose a critical gap in existing agent auditing methodologies. They operate within unconstrained GUIs, interact with arbitrary third-party applications, and rely primarily on visual perception rather than structured environment states. Consequently, standard evaluation signals—such as environment rewards, API-level logs, or deterministic success checks—are often unavailable or unreliable. Given the potential for immediate and costly consequences from misaligned behavior \cite{norman1990automation, lee2004trust}, these characteristics motivate the need for autonomous, scalable, and interface-aware auditing approaches that evaluate CUA behavior directly from observable interactions.

Unlike prior work that evaluates a single auditor or a single platform, our study is the first to systematically analyze cross-platform generalization, confidence calibration, and inter-model disagreement of VLM auditors at scale.

\section{Methodology}

\subsection{Vision-Language Model--Based Auditors}

We study VLMs used as autonomous auditors for evaluating the task completion of CUAs. Given a task instruction and the final GUI state produced by an agent, a VLM auditor is prompted to assess whether the task has been successfully completed. The auditor outputs a binary judgment (\emph{done} or \emph{not done}) together with an associated confidence score.

Formally, for each task instance $i$, the auditor observes a tuple
\[
(x_i, d_i),
\]
where $x_i$ denotes the final screenshot of the GUI environment and $d_i$ is the natural-language task description. The auditor then predicts a probability
\[
p_i^{(m)} \in [0,1],
\]
representing the model’s confidence that the task was successfully completed, where $m$ indexes the auditor model.

And the corresponding predicted done/not done label is defined as:

\[
\hat{y}_i^{(m)} \in \{0,1\},
\]

We evaluate five VLMs as autonomous auditors, spanning both proprietary and open-source families. Among proprietary models, we consider GPT-4o\footnote{\url{https://openai.com/index/hello-gpt-4o/}} and Claude~3.5~Sonnet\footnote{\url{https://claude.com/product/overview}}, selected for their state-of-the-art multimodal perception and reasoning capabilities. For open-source auditors, we evaluate LLaVA-v1.5-7B \cite{liu2024llava15}, InternVL-2-8B \cite{chen2024internvl2}, and Qwen2-VL-7B \cite{bai2024qwen2vl}, which represent strong publicly available alternatives with diverse architectural designs and training regimes.

These models span both proprietary and open-source systems and differ substantially in architecture size, training data, and multimodal reasoning capabilities, enabling a broad analysis of auditor behavior.

\subsection{Benchmarks}

We evaluate VLM auditors using three widely adopted benchmarks for CUAs: Windows Agent Arena, OSWorld, and macOSWorld. Together, these benchmarks cover a diverse set of real-world tasks across major desktop operating systems, including Windows, Linux, and macOS, and span a broad range of applications, interaction patterns, and task complexities.

Each benchmark defines tasks via natural-language instructions and evaluates agent behavior based on task completion in realistic GUI environments. While the underlying environments differ in operating system and application ecosystem, all three benchmarks provide a binary notion of task success, indicating whether a task was successfully completed or not at the end of an episode.

In our study, we adopt this binary \emph{done / not done} task outcome provided by each benchmark as ground-truth supervision. Formally, for each task instance $i$, the benchmark assigns a ground-truth label
\[
y_i \in \{0,1\},
\]
where $y_i = 1$ denotes that the task is deemed \emph{done} by the benchmark’s official evaluation protocol, and $y_i = 0$ denotes \emph{not done}. These labels serve as the reference against which we assess the correctness, calibration, and agreement of VLM-based auditors.

By relying on benchmark-provided success signals rather than human annotations, we ensure scalability and reproducibility of our evaluation while enabling systematic comparison across operating systems and task domains.

\subsection{Calibration and Confidence Assessment}

Beyond binary correctness, we evaluate how well VLM auditors’ confidence scores align with ground-truth task outcomes. Each auditor produces (i) a predicted probability of task success and (ii) a corresponding binary decision. Specifically, for each task instance $i$ and auditor $m$, the model outputs a probability
\[
p_i^{(m)} \in [0,1],
\]
which is thresholded to obtain a predicted label
\[
\hat{y}_i^{(m)} \in \{0,1\},
\]
where $\hat{y}_i^{(m)} = 1$ denotes a prediction of \emph{done} and $\hat{y}_i^{(m)} = 0$ denotes \emph{not done}. The ground-truth label provided by the benchmark is denoted as
\[
y_i \in \{0,1\}.
\]

We measure calibration using the Brier score, a strictly proper scoring rule defined as
\[
\mathrm{Brier}_m = \frac{1}{N} \sum_{i=1}^{N} \left(p_i^{(m)} - y_i\right)^2,
\]
where $N$ is the total number of evaluated tasks.

\[
\mathrm{Std}_m =
\sqrt{
\frac{1}{N}
\sum_{i=1}^{N}
\left(
\left(p_i^{(m)} - y_i\right)^2 - \mathrm{Brier}_m
\right)^2
}.
\]

Since the Brier score is a squared-error metric, lower values correspond to better calibration. Likewise, a lower $\mathrm{Std}_m$ indicates more stable calibration across tasks.

\subsection{Inter-Model Agreement}

Beyond correctness and calibration, we analyze the extent to which different VLM auditors agree in their judgments of task completion. Inter-model agreement captures the consistency of auditing decisions across models and provides insight into task ambiguity and evaluator subjectivity, particularly in settings where success criteria may not be fully observable from the final GUI state.

For each pair of auditors $(m, m')$, we measure agreement on the binary predictions $\hat{y}_i^{(m)} \in \{0,1\}$ using Cohen’s $\kappa$ coefficient. Cohen’s $\kappa$ accounts for agreement occurring by chance and is defined as
\[
\kappa = \frac{p_o - p_e}{1 - p_e},
\]
where $p_o$ denotes the observed agreement rate between two auditors and $p_e$ denotes the expected agreement under independence. Values of $\kappa$ range from $-1$ to $1$, with higher values indicating stronger agreement and $\kappa = 0$ corresponding to chance-level agreement.

We compute pairwise $\kappa$ scores separately for each benchmark and operating system, enabling an analysis of how agreement varies across environments and task distributions. High inter-model agreement suggests that task completion is visually and semantically unambiguous in the final GUI state, whereas low agreement indicates cases where success is difficult to infer, multiple interpretations are plausible, or auditors rely on different implicit assumptions.

By explicitly analyzing inter-model agreement, we move beyond single-model evaluation and characterize the variance and uncertainty inherent in model-based auditing of CUAs.

\section{Results}

In this section, we present evaluation of $5$ VLMs as an auditors of CUA across three operating systems (macOS, Windows and Linux). Our analysis focuses on three complementary aspects: (i) accuracy of task completion assessment, (ii) calibration of confidence estimates, and (iii) inter-model agreement.

\paragraph{Accuracy of Task Completion Assessment.}
Table~\ref{tab:accuracy_all} reports the accuracy of VLM auditors in predicting benchmark-provided \emph{done / not done} labels. Overall, proprietary models outperform open-source alternatives across all benchmarks, with GPT-4o and Claude~3.5~Sonnet achieving the highest accuracy. Performance varies substantially across operating systems: all auditors perform best on macOSWorld, while accuracy drops notably on Windows Agent Arena and OSWorld. This suggests that auditing difficulty is strongly influenced by environment complexity and interaction diversity, rather than by auditor architecture alone.

Among open-source models, InternVL-2-8B and Qwen2-VL-7B consistently outperform LLaVA-v1.5-7B, but still lag behind proprietary models. These results indicate that while open-source VLMs can function as auditors, their reliability remains limited in more complex or heterogeneous environments.

\begin{figure}[t]
    \centering
    \includegraphics[width=\linewidth]{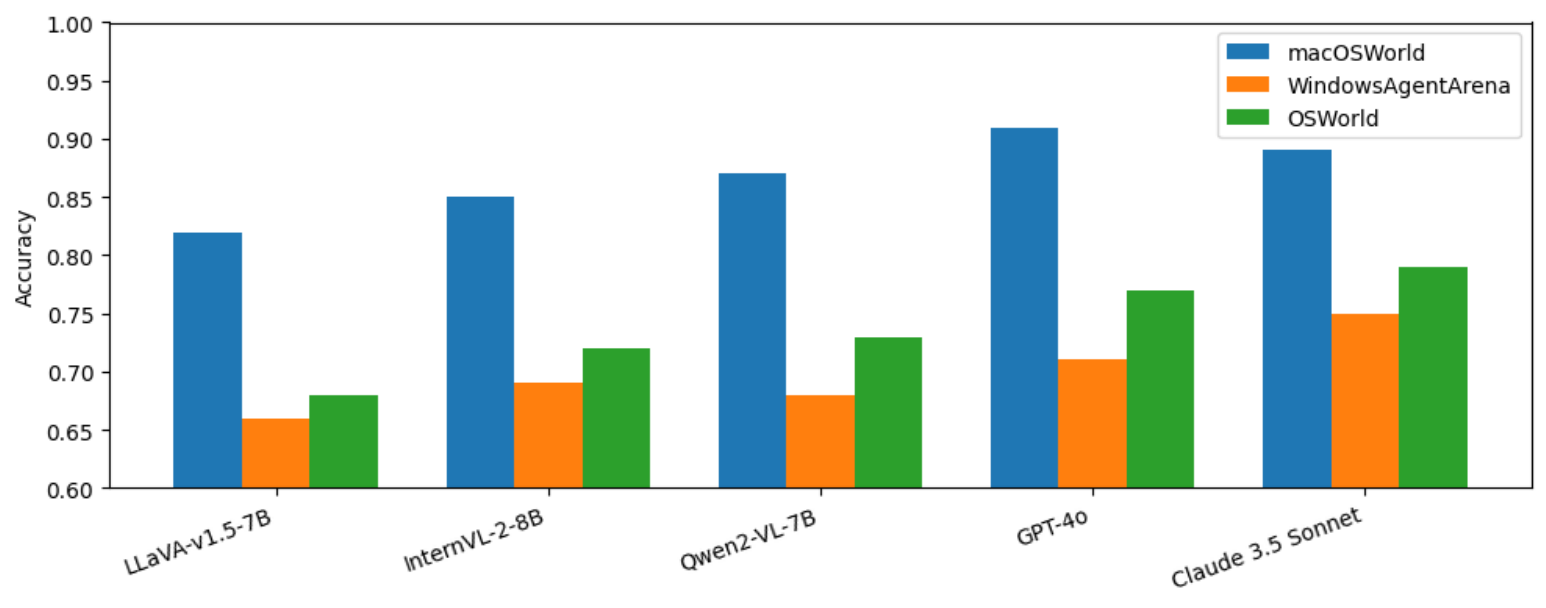}
    \caption{Accuracy of VLM auditors across benchmarks, ordered by increasing mean accuracy across macOSWorld, Windows Agent Arena, and OSWorld.}
    \label{fig:accuracy_auditors}
\end{figure}

\begin{table*}[t]
\centering
\caption{Accuracy of task competion assesment by VLM auditors across benchmarks.}
\label{tab:accuracy_all}
\begin{tabular}{lccc}
\toprule
\textbf{Auditor} & \textbf{macOSWorld} & \textbf{WindowsAgentArena} & \textbf{OSWorld} \\
\midrule
\multicolumn{4}{c}{\textbf{Proprietary Auditors}} \\
\midrule
GPT-4o & \textbf{0.91} & 0.71 & 0.77 \\
Claude 3.5 Sonnet & 0.89 & \textbf{0.75} & \textbf{0.79} \\
\midrule
\multicolumn{4}{c}{\textbf{Open-Source Auditors}} \\
\midrule
InternVL-2-8B & 0.85 & \textbf{0.69} & 0.72 \\
LLaVA-v1.5-7B & 0.82 & 0.66 & 0.68 \\
Qwen2-VL-7B & \textbf{0.87} & 0.68 & \textbf{0.73} \\
\bottomrule
\end{tabular}
\end{table*}

\paragraph{Calibration and Confidence Reliability.}
Beyond accuracy, reliable auditing requires that confidence scores meaningfully reflect uncertainty. Table~\ref{tab:brier_all} reports Brier scores (mean $\pm$ standard deviation) for each auditor, where lower values indicate better calibration. Proprietary models exhibit substantially lower Brier scores across all benchmarks, indicating more reliable confidence estimates. In contrast, open-source models tend to be overconfident or poorly calibrated, particularly on Windows Agent Arena and OSWorld.

Notably, calibration quality does not always track accuracy: some models with comparable accuracy exhibit significantly different Brier scores. This highlights that binary correctness alone is insufficient to characterize auditor reliability, especially in safety-critical or deployment settings where confidence estimates inform downstream decisions.

\begin{table*}[t]
\centering
\caption{Calibration of VLM auditors measured by Brier score (mean $\pm$ std) across benchmarks.}
\label{tab:brier_all}
\begin{tabular}{lccc}
\toprule
\textbf{Auditor} & \textbf{macOSWorld} & \textbf{WindowsAgentArena} & \textbf{OSWorld} \\
\midrule
\multicolumn{4}{c}{\textbf{Proprietary Auditors}} \\
\midrule
GPT-4o & \textbf{$0.058 \pm 0.003$} & \textbf{$0.091 \pm 0.006$} & \textbf{$0.074 \pm 0.004$} \\
Claude 3.5 Sonnet & $0.063 \pm 0.004$ & $0.099 \pm 0.007$ & $0.081 \pm 0.005$ \\
\midrule
\multicolumn{4}{c}{\textbf{Open-Source Auditors}} \\
\midrule
InternVL-2-8B & \textbf{$0.097 \pm 0.007$} &\textbf{ $0.142 \pm 0.010$ }& \textbf{$0.118 \pm 0.008$} \\
LLaVA-v1.5-7B & $0.112 \pm 0.008$ & $0.159 \pm 0.012$ & $0.134 \pm 0.009$ \\
Qwen2-VL-7B & $0.105 \pm 0.008$ & $0.167 \pm 0.011$ & $0.141 \pm 0.010$ \\
\bottomrule
\end{tabular}
\end{table*}

\paragraph{Inter-Model Agreement.}
To assess consistency across auditors, we computed pairwise inter-model agreement using Cohen’s $\kappa$ (Table~\ref{tab:kappa_pairwise}). Agreement is highest between proprietary auditors, indicating relatively consistent judgments in assessing task completion. Agreement between proprietary and open-source models is markedly lower, while agreement among open-source models remains moderate.

Across all auditor pairs, agreement decreases on Windows Agent Arena and OSWorld, suggesting that harder or more ambiguous tasks amplify subjective differences in auditor judgments. These results indicate that even high-performing auditors may disagree substantially in complex environments, underscoring the importance of studying auditor variance rather than relying on a single model.

\begin{table*}[t]
\centering
\caption{Pairwise inter-model agreement of VLM auditors measured using Cohen’s $\kappa$ across benchmarks. Higher is better.}
\label{tab:kappa_pairwise}
\begin{tabular}{llccc}
\toprule
\textbf{Model A} & \textbf{Model B} & \textbf{macOSWorld} & \textbf{WindowsAgentArena} & \textbf{OSWorld} \\
\midrule
\multicolumn{5}{c}{\textbf{Proprietary Auditors}} \\
\midrule
GPT-4o & Claude 3.5 Sonnet & \textbf{0.76} & \textbf{0.66} & \textbf{0.71} \\
\midrule
\multicolumn{5}{c}{\textbf{Proprietary vs Open-Source Auditors}} \\
\midrule
GPT-4o & InternVL-2-8B & 0.64 & 0.57 & 0.61 \\
GPT-4o & LLaVA-v1.5-7B & 0.61 & 0.54 & 0.59 \\
GPT-4o & Qwen2-VL-7B & 0.66 & 0.58 & 0.63 \\
Claude 3.5 Sonnet & InternVL-2-8B & 0.67 & 0.59 & 0.64 \\
Claude 3.5 Sonnet & LLaVA-v1.5-7B & 0.63 & 0.56 & \textbf{0.66} \\
Claude 3.5 Sonnet & Qwen2-VL-7B & \textbf{0.69 }& \textbf{0.61} & 0.6 \\
\midrule
\multicolumn{5}{c}{\textbf{Open-Source Auditors}} \\
\midrule
InternVL-2-8B & LLaVA-v1.5-7B & 0.62 & 0.55 & 0.60 \\
InternVL-2-8B & Qwen2-VL-7B & \textbf{0.68} & 0.60 & \textbf{0.65} \\
LLaVA-v1.5-7B & Qwen2-VL-7B & 0.64 & \textbf{0.67 }& 0.61 \\
\bottomrule
\end{tabular}
\end{table*}

\section{Discussion and Limitations}

Our results indicate that while VLM-based auditing of CUAs is feasible, auditor outputs should be interpreted as uncertain signals rather than definitive judgments. In particular, calibration quality and inter-model agreement provide critical information about auditor reliability that is not captured by accuracy alone. In practical settings, auditor confidence is often used to guide downstream decisions such as whether to request user confirmation, abstain from judgment, or trigger fallback behaviors. Auditors that achieve high accuracy but exhibit poor calibration may therefore still introduce risk by overstating certainty in ambiguous cases.

Inter-model disagreement further highlights the inherent difficulty of inferring task completion from a final GUI state alone. Many CUA tasks depend on hidden system state, background effects, or transient interface changes that may not be visible in a single screenshot. As a result, different auditors may rely on different implicit assumptions when judging success, leading to divergent but individually plausible decisions. Rather than being treated purely as noise, such disagreement can serve as a signal of task ambiguity or insufficient observability, suggesting that additional evidence may be required for reliable evaluation.

This study has several limitations. We restrict auditors to observing only the task instruction and final GUI state, which reflects a scalable and deployment-relevant setting but may underestimate performance for tasks where intermediate actions or temporal context are essential. Our calibration analysis relies on model-reported confidence elicited through standardized prompting, since token-level log probabilities are not consistently accessible across VLMs; consequently, we evaluate the reliability of reported uncertainty rather than intrinsic probabilistic calibration. Finally, we focus exclusively on binary task completion and do not address other important auditing dimensions such as safety, policy compliance, privacy, or harmful side effects, which are critical for real-world deployment of autonomous agents.

\section{Conclusions}

We conducted a large-scale meta-evaluation of VLMs as autonomous auditors for CUAs across three widely used benchmarks spanning macOS, Windows, and Linux. Our results reveal several consistent patterns that have important implications for how model-based evaluation should be designed, reported, and used in practice.

First, auditor performance is strongly environment-dependent. All evaluated models achieve substantially higher accuracy on macOSWorld than on Windows Agent Arena and OSWorld, indicating that auditing difficulty is shaped not only by auditor architecture but also by interface heterogeneity, visual ambiguity, and task diversity across operating systems and applications. As a result, single aggregated performance scores can obscure meaningful failure modes. Reliable auditing therefore requires environment-specific reporting and testing that reflects realistic domain shift rather than averaged metrics alone.

Second, confidence calibration emerges as a critical and independent axis of auditor reliability. Proprietary VLMs exhibit consistently lower Brier scores and more stable confidence estimates, while open-source models are often poorly calibrated, particularly on more challenging benchmarks. Importantly, calibration does not always correlate with accuracy: auditors may make correct judgments while expressing overconfident or unreliable probabilities. This distinction is essential for downstream use, where auditor confidence may guide decisions such as when to request user confirmation, defer execution, or trigger safer fallback policies.

Third, we observe substantial inter-model disagreement, especially on Windows Agent Arena and OSWorld. This disagreement reflects the inherent ambiguity of judging task completion from a final GUI state alone. Many tasks involve hidden state changes, background effects, or success criteria that are not fully observable in a single screenshot, leading different auditors to resolve uncertainty differently. Rather than being treated as noise, disagreement can serve as an informative signal, highlighting ambiguous tasks, implicit benchmark assumptions, or cases where additional evidence beyond the final state is required.

Taken together, these findings suggest concrete implications for both benchmarking and deployment. Benchmarks would benefit from providing richer, verifiable evidence of success—such as structured logs, intermediate states, or checkable artifacts, for tasks where the final GUI state is insufficient. In deployment, oriented evaluation, metrics aligned with safety and reliability, such as calibration quality, robustness under domain shift, and consistency across evaluators, should be prioritized over accuracy alone.

Overall, while VLM-based auditing of CUAs is feasible and proprietary models currently provide the strongest accuracy and calibration, our results show substantial degradation and disagreement in more complex environments. These findings underscore that evaluation itself is a central bottleneck for dependable CUA deployment and must be treated as a first-class research problem, with explicit modeling of evaluator uncertainty, variance, and ambiguity.

\bibliographystyle{ACM-Reference-Format}
\bibliography{references}

\end{document}